\documentclass[journal,twoside,web]{ieeecolor}
\usepackage{generic}
\usepackage{cite}
\usepackage{amsmath,amssymb,amsfonts}
\usepackage{algorithmic}
\usepackage{graphicx}
\usepackage{algorithm,algorithmic}
\usepackage{hyperref}
\hypersetup{hidelinks}
\usepackage{textcomp}
\usepackage{booktabs}
\usepackage{multirow}
\usepackage{threeparttable}
\usepackage{colortbl}
\usepackage{longtable}
\usepackage{rotating}
\usepackage{booktabs, multirow, array}\usepackage{subcaption}
\usepackage{microtype}
\usepackage{amsmath}   
\usepackage{amssymb}   
\usepackage{bbm}
\usepackage{stfloats}

\begin{document}

\title{Understanding the Impact of Model Pruning on Long-Tail Forgetting and Explanation Reliability in Medical Imaging}

\author{Nazish Khalid, Tausifa Jan Saleem, Amal Saqib, Donald C. Wunsch II and Mohammad Yaqub
\thanks{Nazish Khalid and Tausifa Jan Saleem contributed equally to this work.}
\thanks{Nazish Khalid and Donald C. Wunsch are with the Department of Electrical and Computer Engineering, Missouri University of Science and Technology, Rolla, MO 65409 USA
(e-mail: nazishkhalid@mst.edu, dwunsch@mst.edu).}
\thanks{Tausifa Jan Saleem, Amal Saqib and Mohammad Yaqub are with Computing and Mathematical Sciences Division, Mohamed bin Zayed University of Artificial Intelligence, Abu Dhabi, United Arab Emirates (email: tausifa.saleem@mbzuai.ac.ae, amal.saqib@mbzuai.ac.ae, mohammad.yaqub@mbzuai.ac.ae).}
}

\maketitle

\begin{abstract}
Model pruning is widely used to compress deep neural networks, reducing memory and computational requirements with minimal impact on aggregate performance. However, its effect on model behavior remains poorly understood, particularly for long-tailed medical datasets where rare but clinically important conditions are underrepresented. Furthermore, it remains unclear whether pruned models preserve reliable explanations of their predictions. To address this gap, we present a systematic study of long-tail forgetting and explanation reliability under model pruning. Across two long-tailed medical imaging datasets, two CNN architectures, four pruning methods, and sparsity levels up to 95\%, we evaluate predictive performance, explanation stability, and explanation faithfulness. Our results show that predictive performance exhibits a strong frequency-dependent trend, with lower-frequency classes generally experiencing earlier and larger degradation than higher-frequency classes. In contrast, explanation stability and faithfulness are influenced primarily by the pruning strategy, with gradient-informed methods preserving explanation reliability more effectively under aggressive compression. Qualitative and mechanistic analyses further indicate that explanation degradation is primarily associated with the collapse of class-discriminative gradients rather than the disappearance of feature activations. These findings suggest that model compression should be evaluated beyond aggregate performance. Incorporating class-aware and explanation-aware evaluation reveals failure modes that would otherwise remain hidden, while moderate sparsity levels provide a practical balance between compression, predictive performance, and explanation reliability.

\end{abstract}

\begin{IEEEkeywords}
Medical imaging, deep learning, model compression, model pruning, explainability
\end{IEEEkeywords}


\section{Introduction}
\label{sec:intro}

\IEEEPARstart{M}{odel} pruning~\cite{blalock2020state, hoefler2021sparsity, he2023structured, cheng2024survey} has emerged as a principal strategy for reducing the memory footprint and computational cost of deep neural networks. By removing redundant weights, pruning can substantially reduce inference cost with minimal impact on aggregate performance. While such compression is attractive for deploying deep learning models in resource-constrained environments, aggregate performance metrics can mask important changes in model behavior. This issue is especially relevant in medical AI, where failures on clinically important cases may have significant consequences.

Medical imaging datasets are often long-tailed, with common conditions such as pleural effusion and infiltration represented by thousands of training examples, whereas rare but clinically significant conditions such as hernia, pneumomediastinum, and melanoma occur much less frequently~\cite{ju2024monica}. Under such a class imbalance, pruning may disproportionately degrade performance on rare classes. Holste et al.~\cite{holste2023pruning} referred to this phenomenon as \emph{long-tail forgetting}, showing that rare classes are particularly vulnerable to pruning. However, their analysis considered only a single pruning strategy and focused solely on predictive performance.

A second concern is \emph{explanation reliability}. Explainability methods~\cite{guidotti2018survey, arrieta2020explainable} are increasingly embedded in clinical workflows to audit model decisions, identify failure modes, and support clinician trust in automated outputs. If pruning destabilizes these explanations, causing a compressed model to attend to different image regions than its dense counterpart, then a model may retain acceptable aggregate accuracy while producing clinically misleading localizations. This constitutes a \emph{silent safety hazard}: a compressed model that passes standard performance benchmarks yet cannot be trusted to explain its predictions. This concern is particularly important for rare diseases, where model behavior is already difficult to assess due to limited data. The regulatory guidance for AI-based medical systems~\cite{fda2021ai} further emphasizes the importance of explainable decision-making.

Hence, despite the growing interest in model pruning in medical AI~\cite{saleem2026deep}, most studies evaluate pruning using aggregate metrics such as AUC or macro-F1, without class-frequency stratification or assessment of explanation reliability. Consequently, four key questions remain unanswered:

\textbf{Q1}. Is long-tail forgetting inherent to pruning or specific to particular pruning methods?

\textbf{Q2}. How do different pruning strategies affect rare and common classes under comparable sparsity levels?

\textbf{Q3}. How does pruning impact explanation stability and faithfulness?

\textbf{Q4}. How are predictive performance and explanation reliability related under pruning?

To answer these questions, we conduct a systematic study across two long-tailed medical imaging datasets, NIH-CXR-LT~\cite{holste2023pruning} and ISIC-2019~\cite{isic2019}, two architectures (ResNet-50 and DenseNet-121), four pruning strategies (L1-based magnitude pruning~\cite{han2015learning}, SNIP~\cite{lee2018snip}, GraSP~\cite{wang2020picking}, and random pruning), and sparsity levels ranging from 0\% to 95\%. 

The main contributions of this work are:

\begin{enumerate}

\item \textbf{Systematic investigation of long-tail forgetting across pruning strategies (Q1).} We examine whether long-tail forgetting generalizes across different pruning criteria rather than being specific to a particular pruning method.

\item \textbf{Class-frequency-aware comparison of pruning strategies (Q2).} We characterize how different pruning strategies affect classes across the frequency spectrum under comparable sparsity levels.

\item \textbf{Evaluation of explanation reliability under pruning (Q3).} We assess the impact of pruning on explanation reliability along two complementary dimensions: attribution stability and faithfulness.

\item \textbf{Joint analysis of predictive performance and explanation reliability (Q4).} We investigate how predictive performance and explanation reliability evolve and relate to each other under pruning.

\end{enumerate}

To the best of our knowledge, this is the first study to jointly investigate long-tail forgetting and explanation reliability under model pruning in medical imaging.

\section{Related Work}
\label{sec:related}

\subsection{Neural Network Pruning}

Model pruning~\cite{blalock2020state, hoefler2021sparsity, he2023structured, cheng2024survey} reduces the computational and memory requirements of deep neural networks by removing redundant parameters while maintaining predictive performance. Pruning methods differ primarily in how parameter importance is estimated. Early approaches such as Optimal Brain Damage~\cite{lecun1989optimal} and Optimal Brain Surgeon~\cite{hassibi1993optimal} estimate parameter importance using second-order information, whereas magnitude-based pruning~\cite{han2015learning} removes weights with small absolute values and remains a widely adopted baseline because of its simplicity and effectiveness. More recent methods estimate parameter importance using first-order information. SNIP~\cite{lee2018snip} identifies connections based on sensitivity to the training loss at initialization, whereas GraSP~\cite{wang2020picking} preserves gradient flow to improve trainability after pruning. This work compares four representative pruning strategies: magnitude-based pruning (L1), sensitivity-based pruning (SNIP), gradient-preserving pruning (GraSP), and Random pruning.

\subsection{Explainability}

Explainability methods \cite{guidotti2018survey,arrieta2020explainable} provide post-hoc insights into the decision-making process of deep neural networks and are increasingly used to improve transparency and trust in medical AI. Among these, feature attribution methods such as saliency maps \cite{simonyan2013deep}, Integrated Gradients \cite{sundararajan2017axiomatic}, SmoothGrad \cite{smilkov2017smoothgrad}, DeepLIFT \cite{shrikumar2017learning}, and Grad-CAM \cite{selvaraju2017grad} are widely adopted because they highlight image regions that contribute to model predictions. Quantitative evaluation of explanations commonly relies on measures of stability, which assess consistency across models or perturbations, and faithfulness, which evaluates whether highlighted regions causally influence predictions \cite{samek2016evaluating}. These metrics provide complementary perspectives on explanation reliability and have become increasingly important for evaluating trustworthy medical AI systems.

\subsection{Impact of Pruning}

Although pruning often preserves aggregate predictive performance, it can substantially alter class-specific model behavior. Hooker et al.~\cite{hooker2019compressed} showed that pruning disproportionately affects a subset of classes despite preserving overall performance, demonstrating pruning-induced forgetting in natural image classification. Extending this observation to medical imaging, Holste et al.~\cite{holste2023pruning} showed that rare diseases are more vulnerable to pruning than common diseases, identifying the phenomenon of \emph{long-tail forgetting}. More recently, several studies~\cite{merkle2025less,frankle2019dissecting} have investigated the effect of pruning on explainability, reporting that moderate pruning can preserve or even improve explanation quality, whereas aggressive compression often degrades attribution fidelity.

Despite these advances, two important gaps remain. First, existing medical studies of long-tail forgetting have primarily focused on magnitude-based pruning, leaving it unclear whether the phenomenon generalizes across different pruning strategies. Second, prior work typically evaluates predictive performance or explainability independently, without jointly analyzing how compression affects both. To address these gaps, we conduct a comprehensive empirical study of four representative pruning strategies (L1, SNIP, GraSP, and Random) across two long-tailed medical imaging datasets and two widely used CNN architectures. Beyond predictive performance, we quantify explanation reliability using attribution stability and faithfulness, providing a unified analysis of how pruning influences both long-tail forgetting and explanation reliability.
\begin{figure*}[!b]
    \centering
    \includegraphics[
        width=\textwidth,
        height=0.8\textheight,
        keepaspectratio
    ]{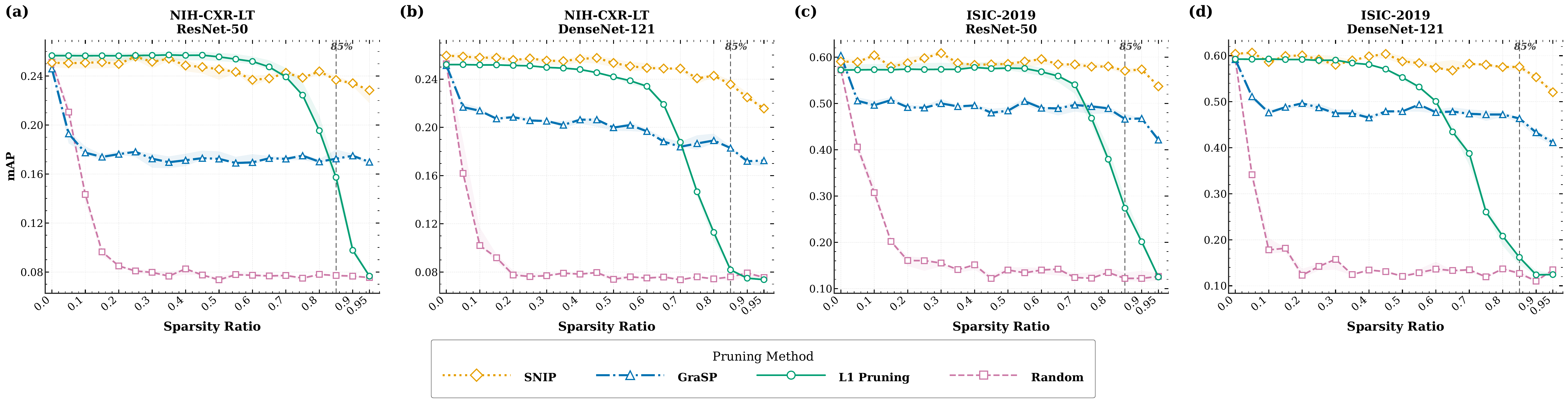}
\caption{mAP as a function of sparsity. Each curve shows the median mAP over three random seeds. SNIP exhibits the most graceful performance degradation, whereas L1 pruning undergoes a cliff-edge collapse at high sparsity. In contrast, Random pruning degrades rapidly even at low sparsity. Panels correspond to (a) NIH-CXR-LT/ResNet-50, (b) NIH-CXR-LT/DenseNet-121, (c) ISIC-2019/ResNet-50, and (d) ISIC-2019/DenseNet-121.}
    \label{fig:map_sparsity}
\end{figure*}
\begin{figure*}[!htb]
    \centering
    \includegraphics[width=\textwidth]{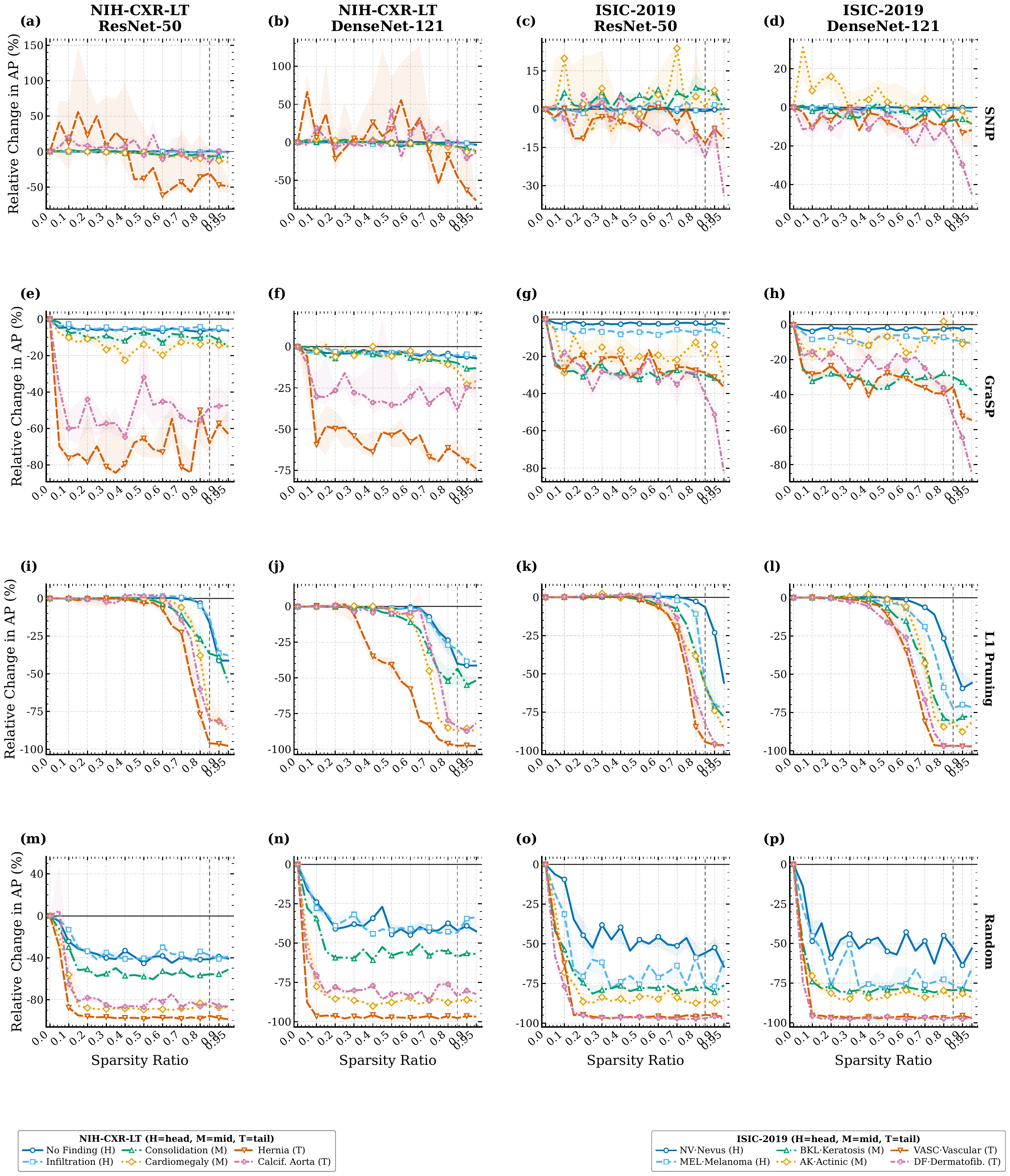}
    \caption{Relative change in per-class AP as a function of sparsity for representative head, mid, and tail classes. Lower-frequency classes generally experience earlier and larger performance degradation than higher-frequency classes. SNIP provides the best preservation of class-wise performance, GraSP exhibits an early performance drop followed by relatively stable degradation, L1 pruning undergoes a pronounced high-sparsity collapse, and Random pruning degrades rapidly across all class-frequency groups. Panels correspond to (a--d) SNIP, (e--h) GraSP, (i--l) L1 pruning, and (m--p) Random pruning, with datasets/backbones ordered as NIH-CXR-LT/ResNet-50, NIH-CXR-LT/DenseNet-121, ISIC-2019/ResNet-50, and ISIC-2019/DenseNet-121.}
    \label{fig:perclass_forgetting}
\end{figure*}
\section{Experimental Setup}

\subsection{Datasets}

We conduct experimentation on two publicly available long-tailed medical imaging datasets spanning clinically distinct domains, thoracic radiology and dermatology, providing complementary settings for evaluating pruning behavior across different imaging modalities, task formulations, and degrees of class imbalance. Table~\ref{tab:datasets} summarizes their key characteristics, including dataset size, class distribution, and long-tail statistics.

\subsubsection{NIH-CXR-LT}

NIH-CXR-LT~\cite{holste2023pruning} is a long-tailed extension of the ChestX-ray14 dataset~\cite{wang2017chestx}, comprising 20 classes (19 thoracic diseases and \textit{No Finding}). We follow the official training, validation, and test splits provided with the benchmark. Following~\cite{holste2023pruning}, classes are grouped into head, mid, and tail categories according to their training frequency.

\subsubsection{ISIC-2019}

ISIC-2019~\cite{isic2019} contains dermoscopic images from eight skin-lesion categories with a pronounced long-tailed class distribution. We exclude the \textit{Unknown} test class and adopt an 85\%/15\% stratified training/validation split while evaluating on the official challenge test set. To ensure a consistent experimental setup across datasets, we formulate ISIC-2019 as a one-vs-rest classification problem using sigmoid outputs and binary cross-entropy loss, matching the multi-label formulation used for NIH-CXR-LT.
\begin{table}[!t]
\centering
\caption{Characteristics of NIH-CXR-LT and ISIC-2019: modality, task, data splits, class imbalance, and head/mid/tail tier composition (tiers defined by training-set frequency).}
\label{tab:datasets}
\footnotesize
\setlength{\tabcolsep}{3pt}
\renewcommand{\arraystretch}{1.05}
\begin{threeparttable}
\resizebox{\columnwidth}{!}{%
\begin{tabular}{@{}lcc@{}}
\toprule
\textbf{Characteristic} & \textbf{NIH-CXR-LT~\cite{holste2023pruning}} & \textbf{ISIC-2019~\cite{isic2019}} \\
\midrule
Modality               & Chest X-ray          & Dermoscopy       \\
Task / classes         & Multi-label / 20     & Single-label / 8 \\
\midrule
Train / Val / Test     & 78,506 / 12,533 / 21,081 & 21,532 / 3,799 / 8,238 \\
\midrule
Most freq.\ (train / test)  & No Finding (44,625 / 8,015) & NV\tnote{a} (10,944 / $\sim$2,454) \\
Least freq.\ (train / test) & Pneumomediast.\ (88) / Calc.\ Aorta (55) & DF\tnote{b} (203 / $\sim$128) \\
Imb.\ ratio (train / test)  & \textbf{507:1} / 146:1 & \textbf{53.9:1} / 19.2:1 \\
\midrule
Head / Mid / Tail\tnote{c}  & 13 / 3 / 4 & 4 / 2 / 2 \\
\bottomrule
\end{tabular}}
\begin{tablenotes}[flushleft]
\scriptsize
\item[a] NV = Melanocytic Nevus. \item[b] DF = Dermatofibroma.
\item[c] Head $>$1,000; Mid 500--1,000; Tail $<$500 training images.
\end{tablenotes}
\end{threeparttable}
\end{table}

\subsection{Models}

We evaluate two widely used CNN backbones: ResNet-50~\cite{he2016deep} and DenseNet-121~\cite{huang2017densely}. These architectures represent two established CNN design paradigms and differ substantially in connectivity (residual versus dense) and model capacity (approximately 25.6M versus 8.0M parameters), providing complementary settings for assessing whether pruning-induced behavior generalizes across CNN architectures. Both models are initialized with ImageNet-pretrained weights and fine-tuned end-to-end.

For each backbone, the original classification layer is replaced with a task-specific linear layer, producing 20 outputs for NIH-CXR-LT and 8 outputs for ISIC-2019. Following the one-vs-rest formulation adopted in this work, sigmoid activations are applied independently to each output, and the networks are optimized using binary cross-entropy with logits. No class reweighting or resampling is employed, preserving the natural long-tailed class distribution throughout training.

\subsection{Pruning Strategies}

We compare four representative one-shot unstructured pruning strategies spanning magnitude-based, gradient-based, and random pruning: L1-based magnitude pruning, SNIP, GraSP, and Random pruning. We focus on these methods to compare distinct weight-importance criteria. For all methods, sparsity levels of 0\%, 5\%, 10\%, 15\% \ldots, 95\% are evaluated, with 0\% corresponding to the dense baseline. SNIP and GraSP compute their masks on the ImageNet-pretrained model, after which the sparse network is fine-tuned. L1 and Random pruning are applied one-shot to the trained dense baselines without further retraining, following~\cite{holste2023pruning}. Differences between the two families therefore reflect both the pruning criterion and whether training follows pruning.

Let $\theta_i$ denote a network weight, $m_i \in \{0,1\}$ the corresponding pruning mask, and $\mathcal{L}$ the training loss.

\subsubsection{L1-based Magnitude Pruning}

Magnitude-based pruning~\cite{han2015learning} removes weights with the smallest absolute values under the assumption that low-magnitude parameters contribute least to the model output. The pruning mask is defined as

\begin{equation}
m_i = \mathbbm{1}\!\left[|\theta_i| \geq \tau_s\right],
\end{equation}

where $\tau_s$ is the global threshold selected to achieve sparsity level $s$.

\subsubsection{SNIP}

SNIP~\cite{lee2018snip} estimates the importance of each weight using its first-order sensitivity to the training loss at initialization. The saliency score is defined as

\begin{equation}
\phi_i =
\left|
\theta_i
\frac{\partial \mathcal{L}}
{\partial \theta_i}
\right|,
\end{equation}

where $\phi_i$ quantifies the contribution of weight $\theta_i$ to the training loss. Weights with the smallest saliency scores are removed. By preserving the most loss-sensitive connections, SNIP seeks to maintain the network's optimization capability under high sparsity.
\begin{figure*}[!t]
    \centering
    \includegraphics[width=\textwidth,
]{
        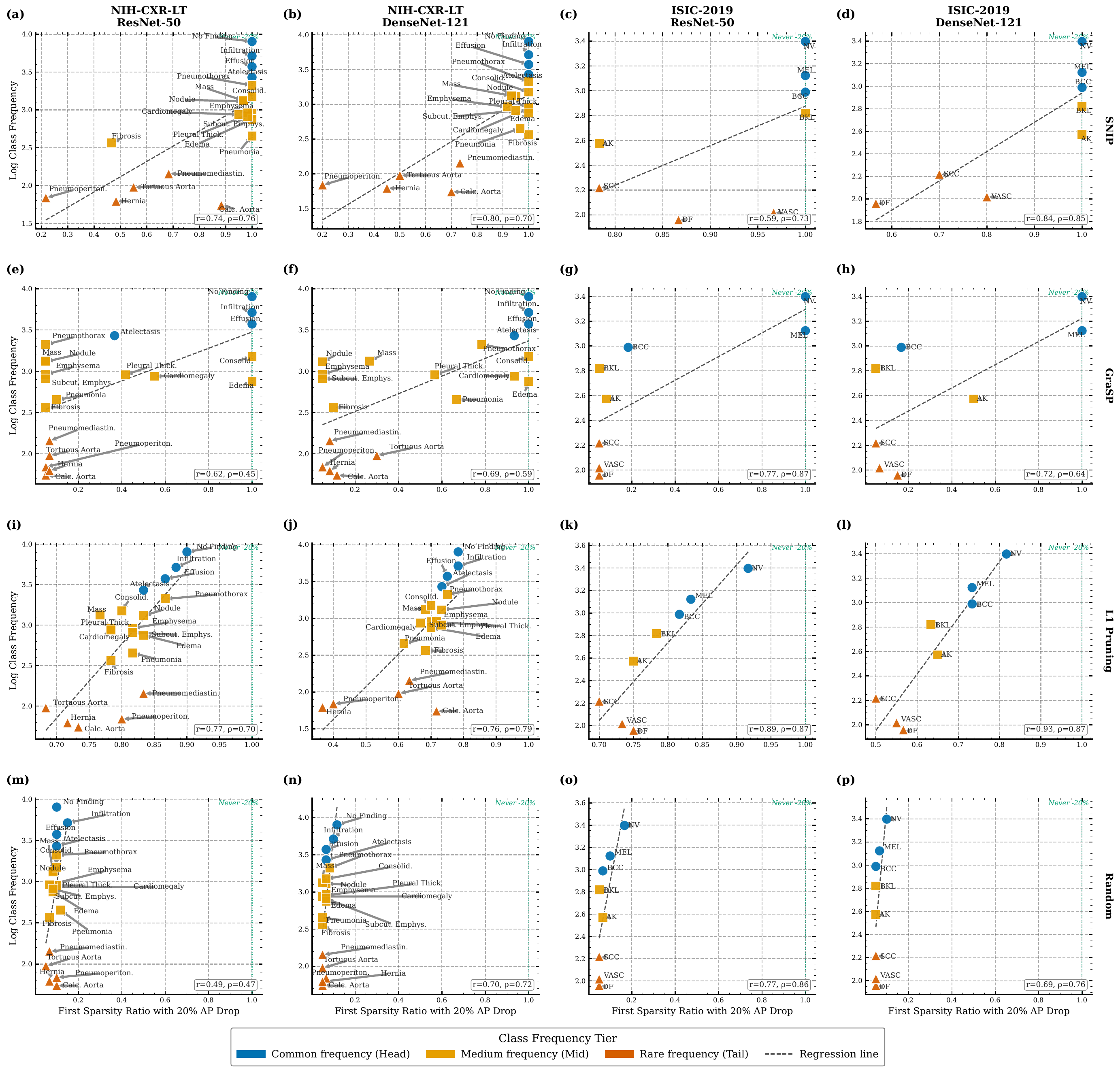}

\caption{First sparsity ratio at which each class experiences a $\geq$20\% reduction in AP versus the logarithm of class frequency. Pearson ($r$) and Spearman ($\rho$) coefficients quantify the association between class frequency and pruning robustness. Positive correlations are observed across pruning methods, indicating that higher-frequency classes generally tolerate greater sparsity before experiencing substantial performance degradation, although the strength of this relationship varies by pruning strategy.}
    \label{fig:scatter_drop}
\end{figure*}
\begin{figure*}[!htb]
    \centering
    \includegraphics[width=\textwidth,
]{
       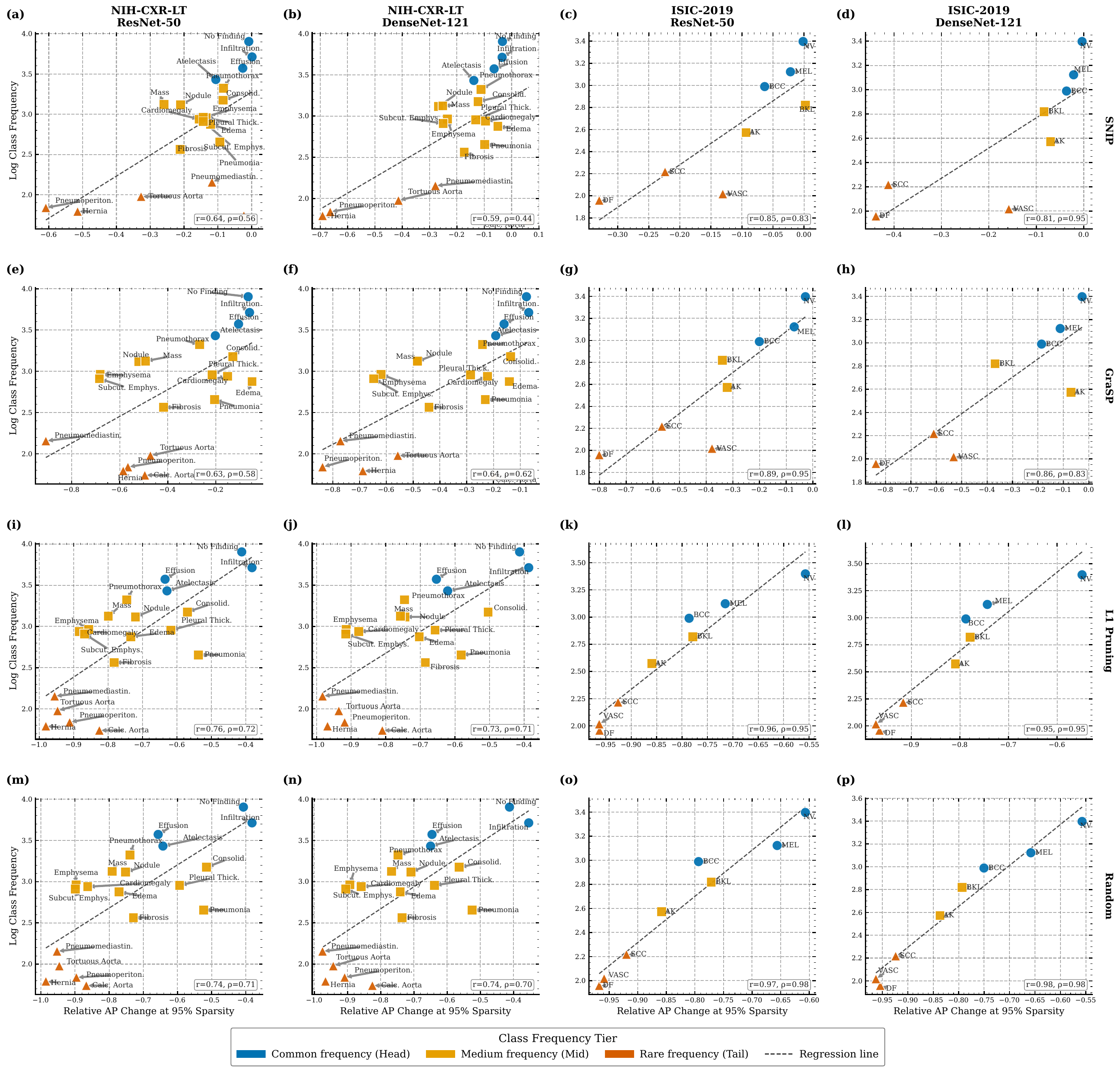}

\caption{Relative change in per-class AP at 95\% sparsity versus the logarithm of class frequency. $r$ and $\rho$ quantify the relationship between class frequency and pruning-induced performance degradation. More severe performance degradation is generally observed for lower-frequency classes, reinforcing the frequency-dependent nature of pruning-induced long-tail forgetting.}
    \label{fig:scatter_rel}
\end{figure*}
\subsubsection{GraSP}

GraSP~\cite{wang2020picking} estimates parameter importance by preserving gradient flow during training using second-order information. Its saliency score is computed as

\begin{equation}
\phi_i = -\left(\mathbf{H}\mathbf{g}\right)_i \cdot \theta_i,
\end{equation}

where $\mathbf{g}=\nabla_{\theta}\mathcal{L}$ is the gradient and $\mathbf{H}=\nabla_{\theta}^{2}\mathcal{L}$ is the Hessian of the loss. By accounting for the effect of pruning on optimization dynamics, GraSP seeks to retain weights that best preserve gradient flow after pruning.

\subsubsection{Random Pruning}

Random pruning removes weights uniformly at random without considering parameter magnitude or gradient information. It serves as a lower-bound baseline for evaluating the effectiveness of importance-based pruning strategies.

\subsection{Training Protocol}
\label{subsec:training}

Models are optimized using Adam with a learning rate of $1\times10^{-4}$ for up to 60 epochs, with early stopping (patience = 15) based on validation performance. An effective batch size of 256 is used throughout. Input images are resized to $224\times224$, normalized using ImageNet statistics, and augmented during training using random horizontal flips and random rotations of $\pm10^\circ$. All models are initialized with ImageNet-pretrained weights and implemented in PyTorch, with Grad-CAM explanations generated using the \texttt{Captum} library. Each experiment is repeated over three random seeds, resulting in a total of 960 evaluated experimental configurations. All experiments are run on NVIDIA V100 and H100 GPUs on the Mill HPC cluster~\cite{mill2024}.

\section{Evaluation Framework}
\label{sec:xai}

We evaluate how model pruning influences both predictive performance and the reliability of post-hoc explanations, with particular emphasis on long-tailed class behavior. Explanation reliability is assessed using Grad-CAM~\cite{selvaraju2017grad} along two complementary dimensions: \textit{explanation stability}, which measures consistency with the dense model, and \textit{explanation faithfulness}, which evaluates whether highlighted regions causally contribute to the model's predictions. Together, these metrics provide a comprehensive assessment of how pruning affects both predictive behavior and explanation reliability.
\begin{figure*}[!htb]
    \centering
    \includegraphics[width=\textwidth]{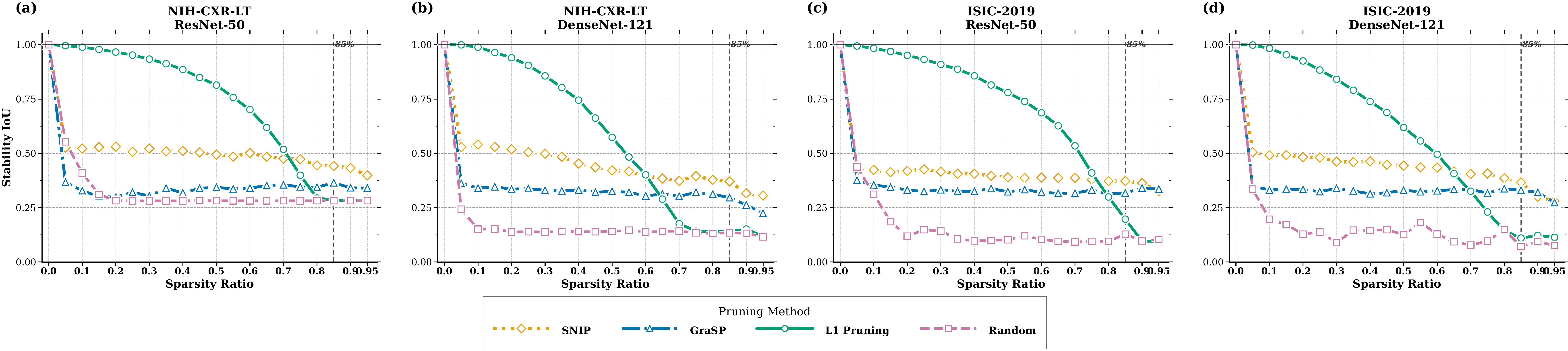}
   \caption{Explanation stability, measured using Stability IoU, as a function of sparsity. Higher Stability IoU indicates greater spatial agreement between Grad-CAM explanations of the dense and pruned models. L1 pruning preserves explanation stability at low-to-moderate sparsity but undergoes a cliff-edge collapse at high sparsity. SNIP exhibits the most gradual degradation, whereas GraSP undergoes an early reduction followed by a relatively stable plateau and Random pruning rapidly falls to and remains at low Stability IoU.}
    \label{fig:stability_1x4}
\end{figure*}
\begin{figure*}[!htb]
    \centering
    \includegraphics[width=\textwidth]{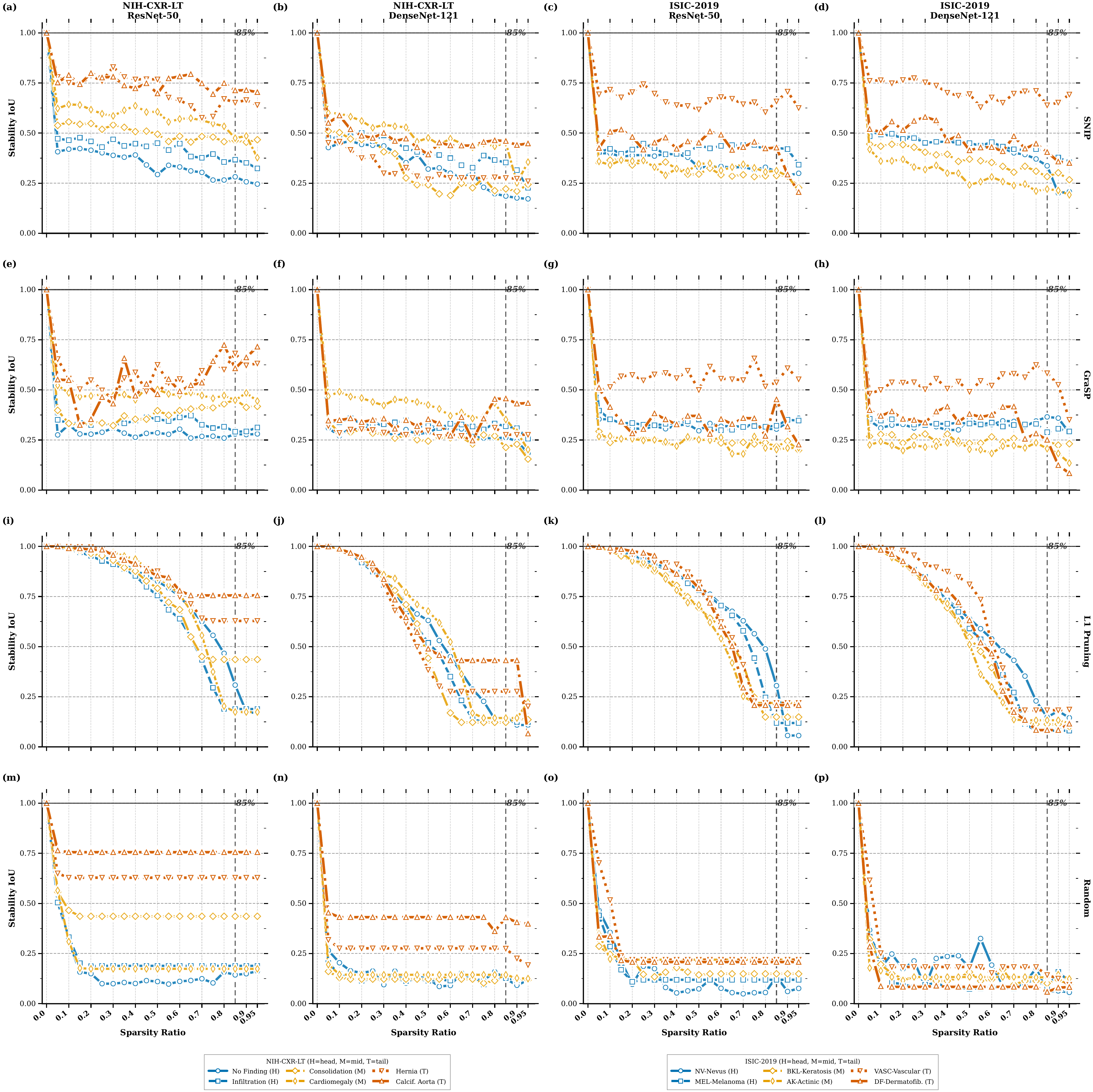}
\caption{Per-class explanation stability (Stability IoU) as a function of sparsity for representative head, mid, and tail classes. Explanation stability is primarily influenced by the pruning strategy rather than by class frequency, with no consistent ordering across head, mid, and tail classes.}

    \label{fig:stability_4x4}
\end{figure*}
\subsection{Predictive Performance Metrics}

Predictive performance is evaluated using mean Average Precision (mAP) and per-class Average Precision (AP). For class $c$,

\begin{equation}
\text{AP}_c =
\sum_{k=1}^{K}
\left(
R_c^{(k)}-R_c^{(k-1)}
\right)
P_c^{(k)},
\end{equation}

and

\begin{equation}
\text{mAP}
=
\frac{1}{C}
\sum_{c=1}^{C}
\text{AP}_c,
\end{equation}

where $P_c^{(k)}$ and $R_c^{(k)}$ denote precision and recall at the $k$-th operating point, $K$ is the number of thresholds, and $C$ is the number of classes.

AP is preferred over AUROC because it is more sensitive to severe class imbalance and better reflects performance on rare classes. Per-class AP is further stratified by class frequency to quantify long-tail forgetting.

\subsection{Explanation Reliability}

\subsubsection{Attribution Method}

For an input image $\mathbf{x}$ and target class $c$, Grad-CAM generates the attribution map

\begin{equation}
\mathbf{A}^{(c)}
=
\mathrm{ReLU}
\left(
\sum_k
\alpha_k^{(c)}
\mathbf{F}^k
\right),
\qquad
\alpha_k^{(c)}
=
\frac{1}{HW}
\sum_{i,j}
\frac{\partial z_c}
{\partial F_{ij}^{k}},
\end{equation}

where $\mathbf{F}^k$ denotes the $k$-th feature map, $\alpha_k^{(c)}$ its corresponding importance weight, and $z_c$ the class score.

\subsubsection{Explanation Stability}

Explanation stability quantifies how pruning alters Grad-CAM explanations relative to the corresponding dense model. Stability is measured using Stability IoU, computed as the Intersection-over-Union (IoU) between the top 20\% most salient pixels in the dense and pruned attribution maps:

\begin{equation}
\text{Stability IoU}
=
\frac{
|S_{\mathrm{dense}}
\cap
S_{\mathrm{pruned}}|
}{
|S_{\mathrm{dense}}
\cup
S_{\mathrm{pruned}}|
}.
\end{equation}

Here, $S_{\mathrm{dense}}$ and $S_{\mathrm{pruned}}$ denote the sets of top 20\% salient pixels from the dense and pruned Grad-CAM maps, respectively. Stability IoU is averaged over all test images and positive class labels.

\subsubsection{Explanation Faithfulness}

Explanation faithfulness is evaluated using the Area Over the Perturbation Curve (AOPC), which measures the reduction in prediction confidence as progressively more salient image regions are masked:

\begin{equation}
\mathrm{AOPC}
=
\frac{1}{T}
\sum_{t=1}^{T}
\frac{
p_c^{(0)}
-
p_c^{(t)}
}{
p_c^{(0)}
},
\end{equation}

where $p_c^{(0)}$ is the original prediction probability, $p_c^{(t)}$ is the probability after the $t$-th perturbation step, and $T=20$. Higher AOPC values indicate more faithful explanations. Samples with $p_c^{(0)}<0.01$ are excluded.

\subsubsection{Evaluation Protocol}

Explainability metrics are computed on stratified subsets of 500 test images per dataset, preserving the original class-frequency distribution while keeping the evaluation computationally tractable. Grad-CAM maps are generated for all positive class labels, normalized to the range $[0,1]$, and used to compute Stability IoU and AOPC. Stability IoU is evaluated relative to the corresponding dense model, whereas AOPC is computed independently for each model. All reported explainability metrics are averaged over three random seeds.

\section{Results}
\label{sec:experiments}
\subsection{Impact of Pruning on Predictive Performance}
\label{sec:results_performance}

\begin{figure*}[!htb]
    \centering
    \includegraphics[width=\textwidth]{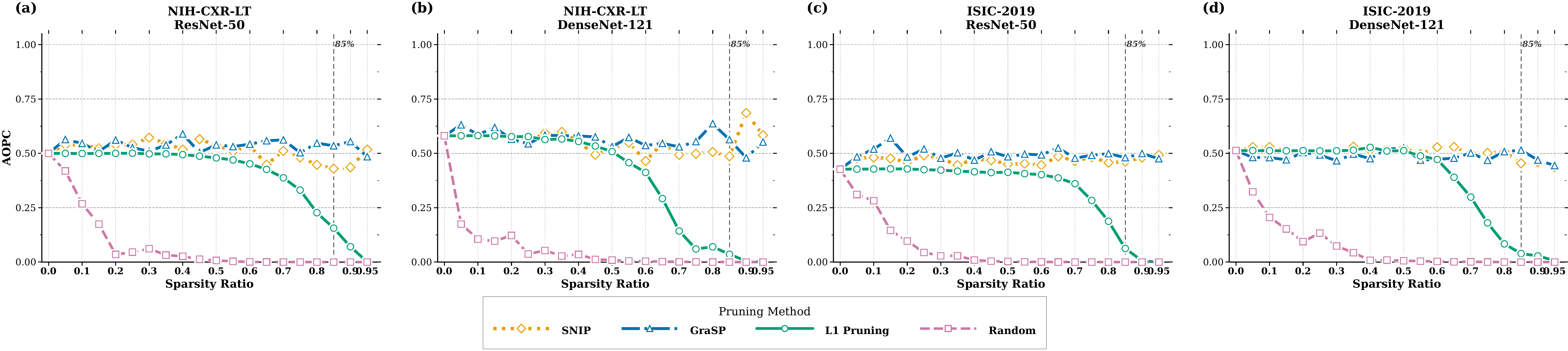}
    \caption{Explanation faithfulness, measured using AOPC, as a function of sparsity. Higher AOPC indicates greater explanation faithfulness, meaning that the highlighted image regions contribute more strongly to the model's prediction. SNIP and GraSP preserve explanation faithfulness across a wide range of sparsity levels, whereas L1 pruning exhibits a sharp collapse at high sparsity and Random pruning degrades rapidly at low sparsity.}
    \label{fig:aopc_1x4}
\end{figure*}
\begin{figure*}[!t]
    \centering
    \includegraphics[width=\textwidth]{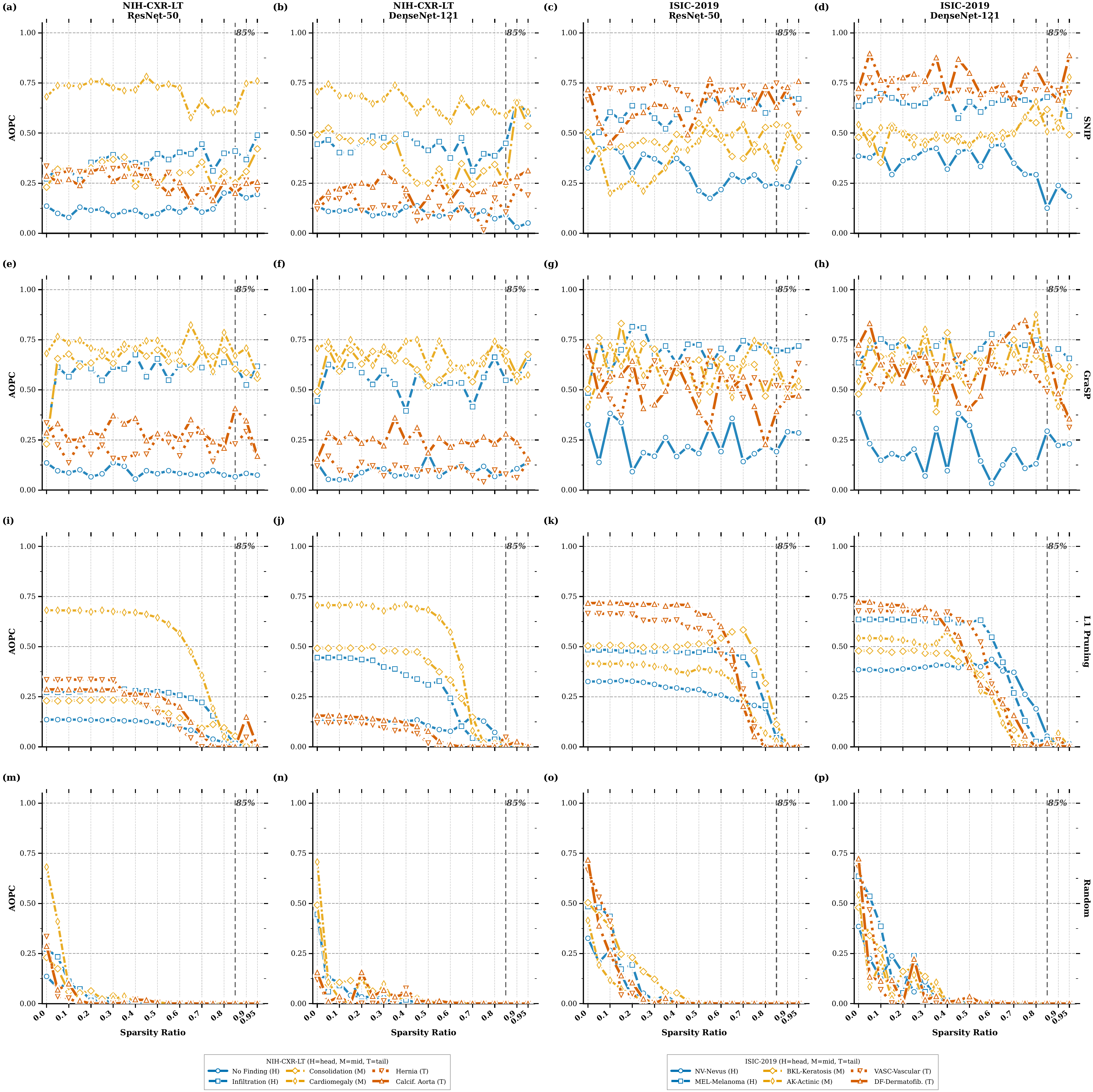}
    \caption{Per-class AOPC as a function of sparsity for representative head, mid, and tail classes. Explanation faithfulness is primarily influenced by the pruning strategy, with no consistent head--mid--tail ordering observed across representative classes.}
    \label{fig:aopc_4x4}
\end{figure*}
\begin{figure*}[t]
    \centering

    \begin{subfigure}[t]{0.48\textwidth}
        \centering
        \includegraphics[width=\linewidth]{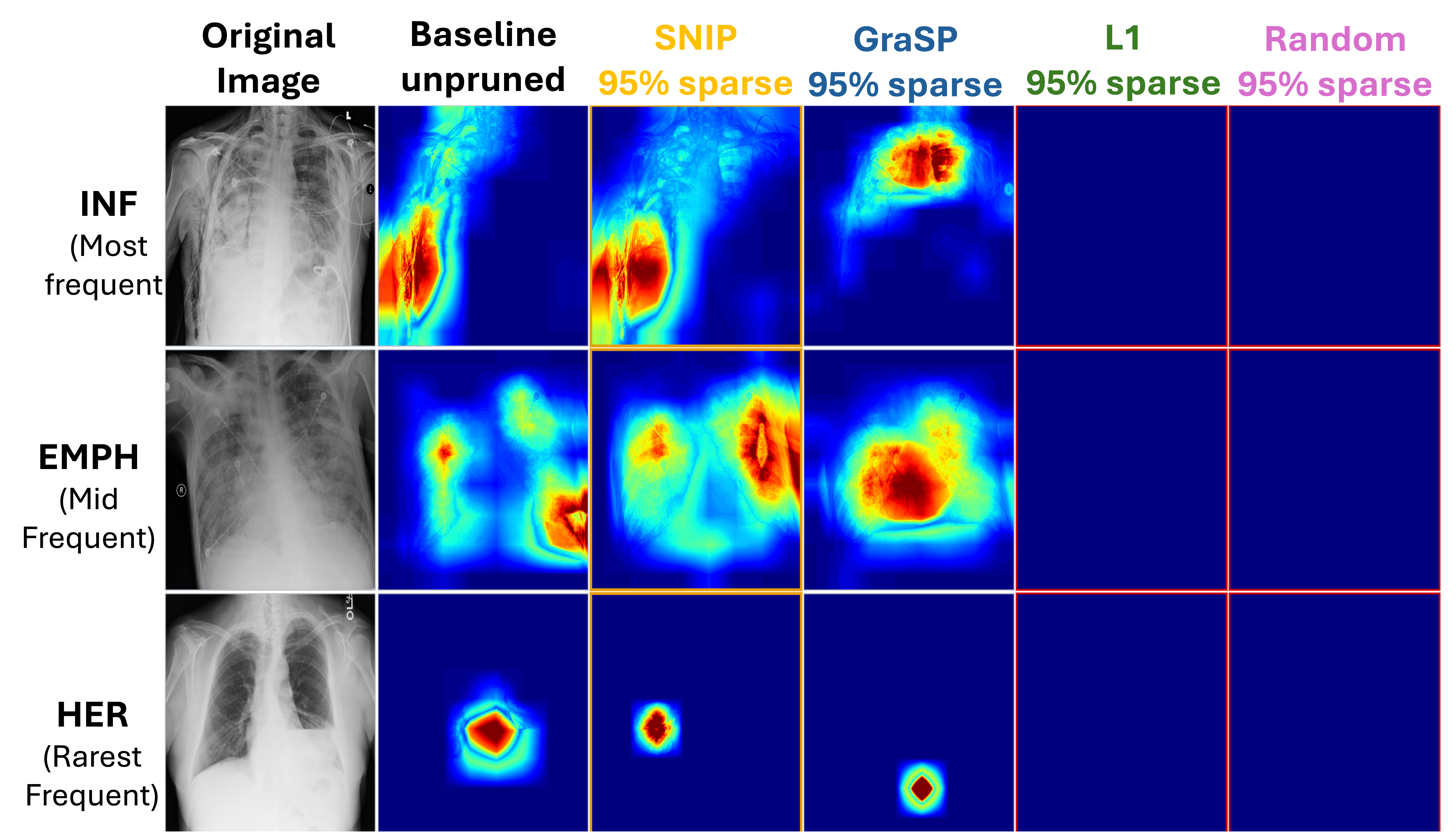}
        \caption{NIH-CXR-LT/ResNet-50}
        \label{fig:nih_resnet50_xai}
    \end{subfigure}
    \hfill
    \begin{subfigure}[t]{0.48\textwidth}
        \centering
        \includegraphics[width=\linewidth]{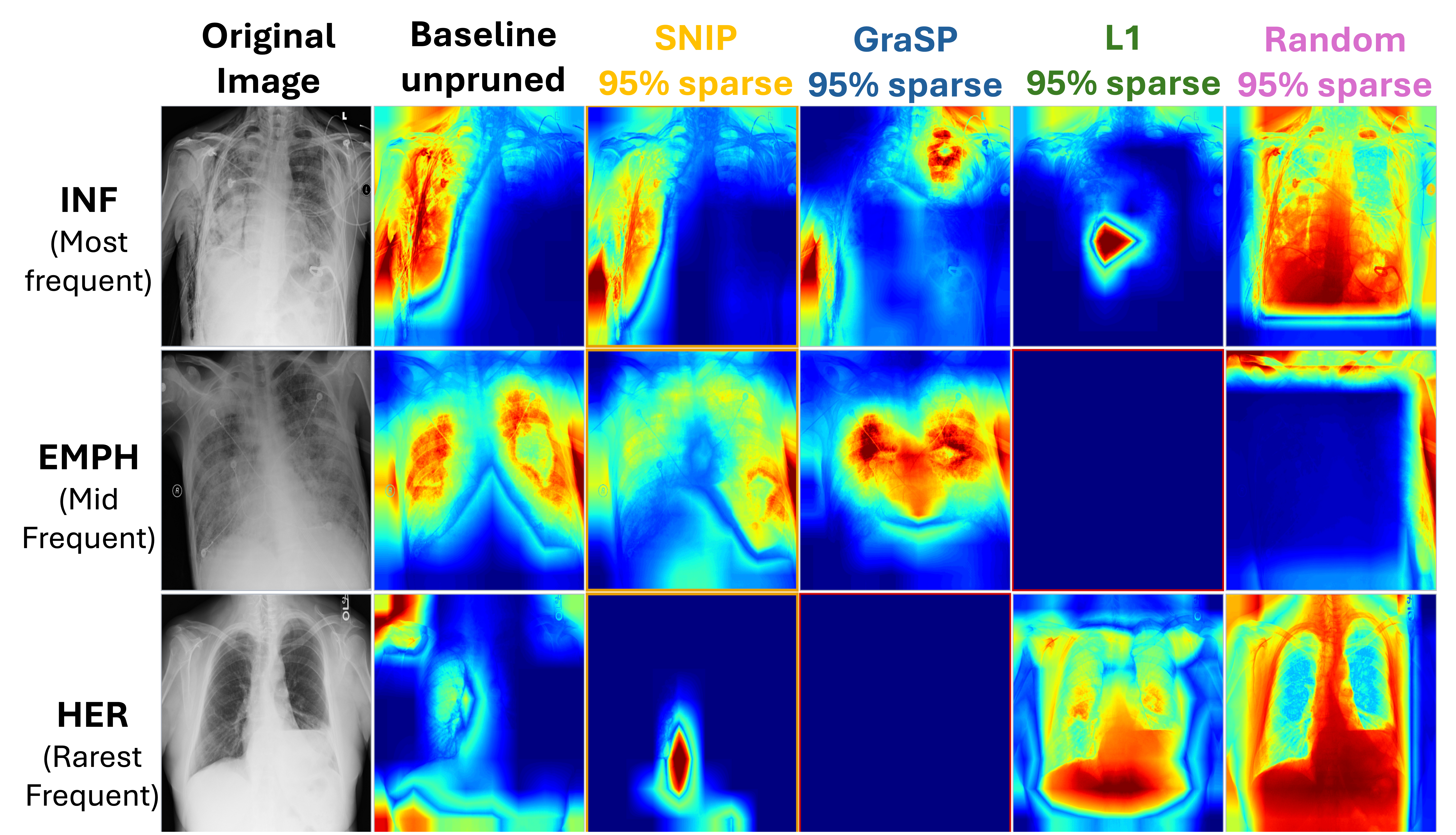}
        \caption{NIH-CXR-LT/DenseNet-121}
        \label{fig:nih_densenet121_xai}
    \end{subfigure}

    \vspace{0.6em}

    \begin{subfigure}[t]{0.48\textwidth}
        \centering
        \includegraphics[width=\linewidth]{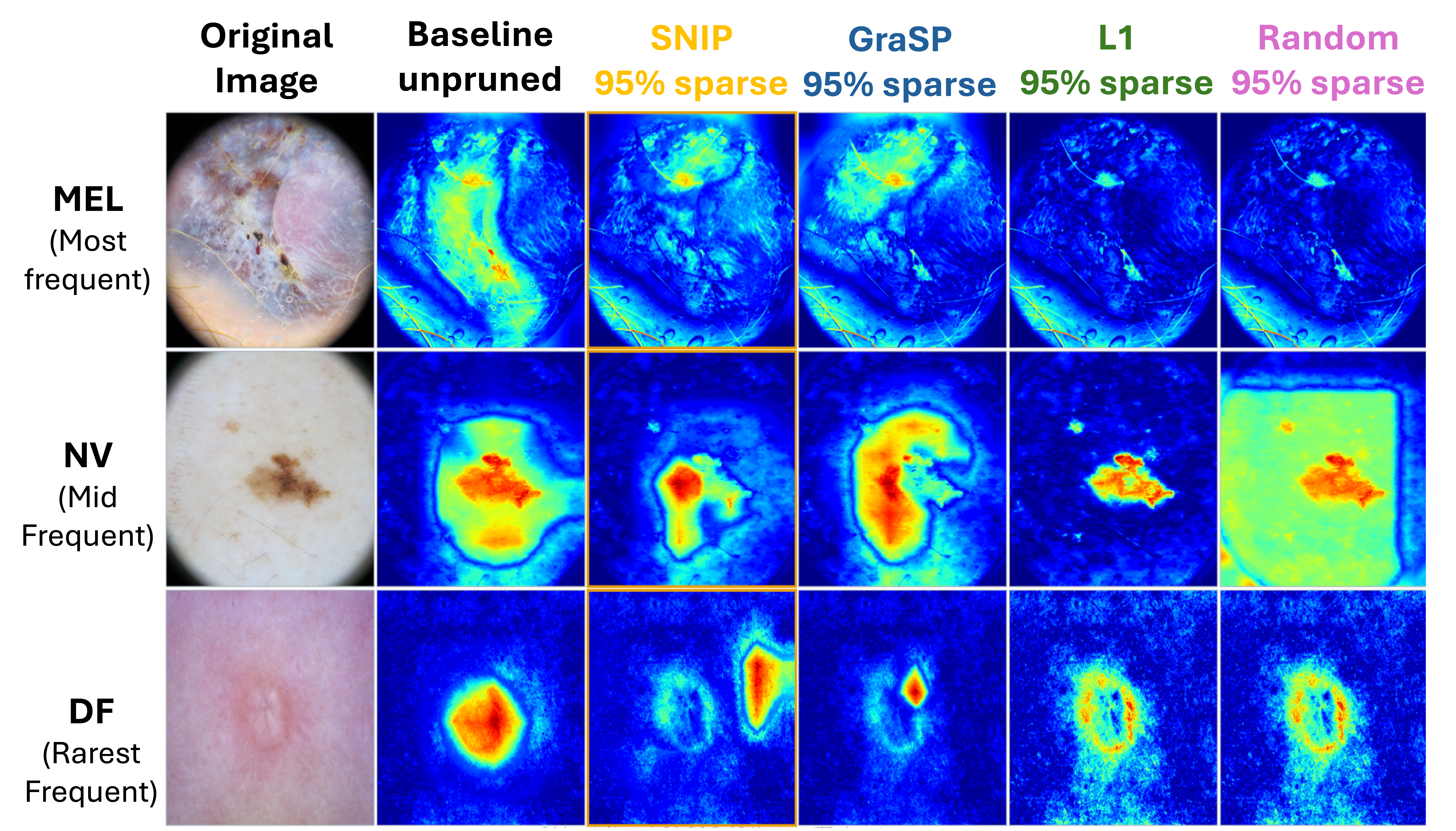}
        \caption{ISIC-2019/ResNet-50}
        \label{fig:isic_resnet50_xai}
    \end{subfigure}
    \hfill
    \begin{subfigure}[t]{0.48\textwidth}
        \centering
        \includegraphics[width=\linewidth]{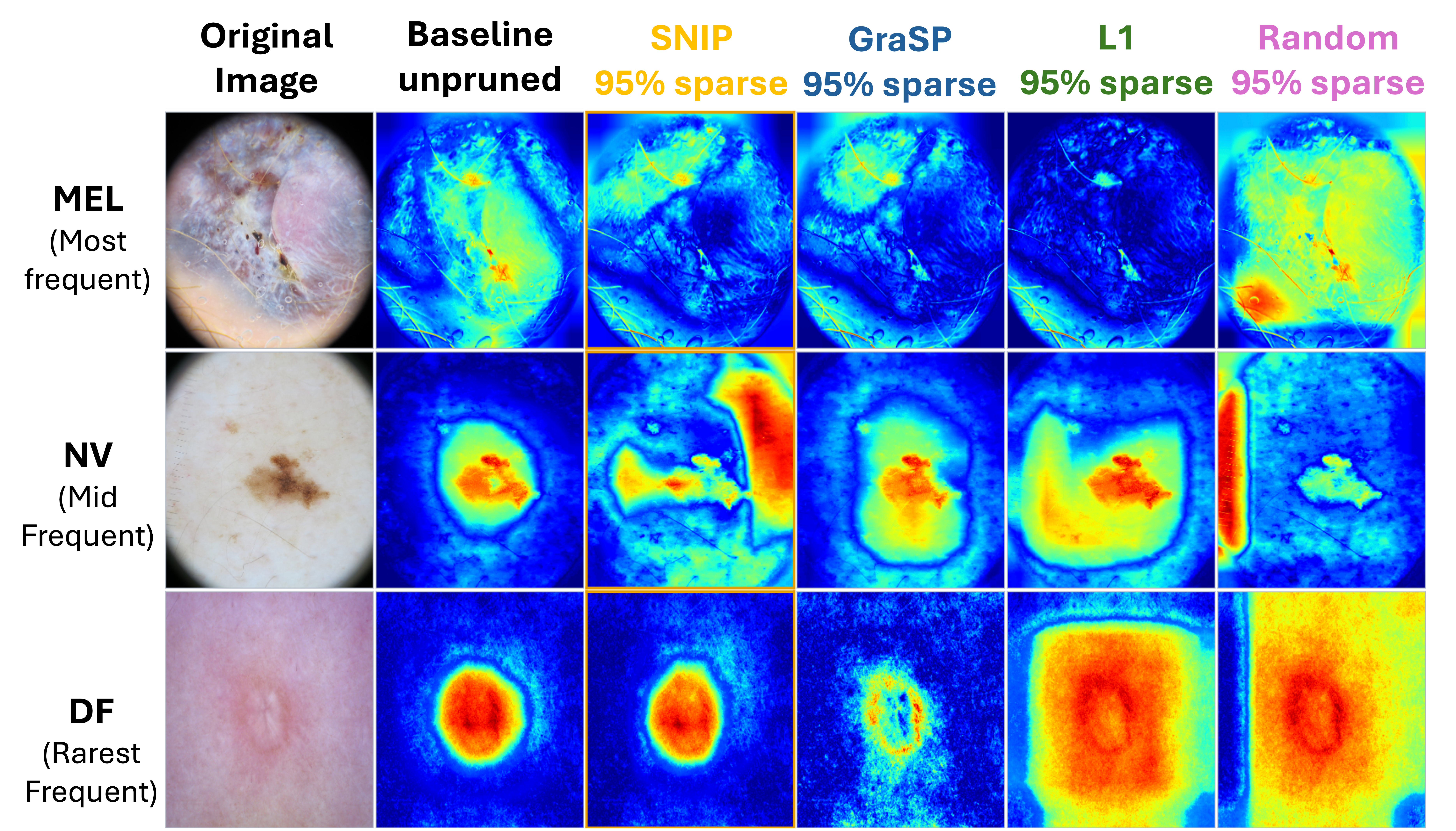}
        \caption{ISIC-2019/DenseNet-121}
        \label{fig:isic_densenet121_xai}
    \end{subfigure}

\caption{Qualitative Grad-CAM comparison under different pruning methods at 95\% sparsity. Each panel fixes the dataset and backbone, comparing the unpruned baseline against SNIP, GraSP, L1 pruning, and Random pruning. Representative head, mid, and tail classes are shown for each dataset: infiltration (INF), emphysema (EMPH), and hernia (HER) for NIH-CXR-LT, and melanoma (MEL), melanocytic nevi (NV), and dermatofibroma (DF) for ISIC-2019. SNIP most consistently preserves lesion-centered localization relative to the unpruned baseline, GraSP generally retains meaningful activation regions, whereas L1 and Random pruning frequently produce degraded or implausible explanations at 95\% sparsity.}
    \label{fig:xai_pruning_comparison}
\end{figure*}
\begin{figure}[!htb]
    \centering
    \includegraphics[width=\linewidth]{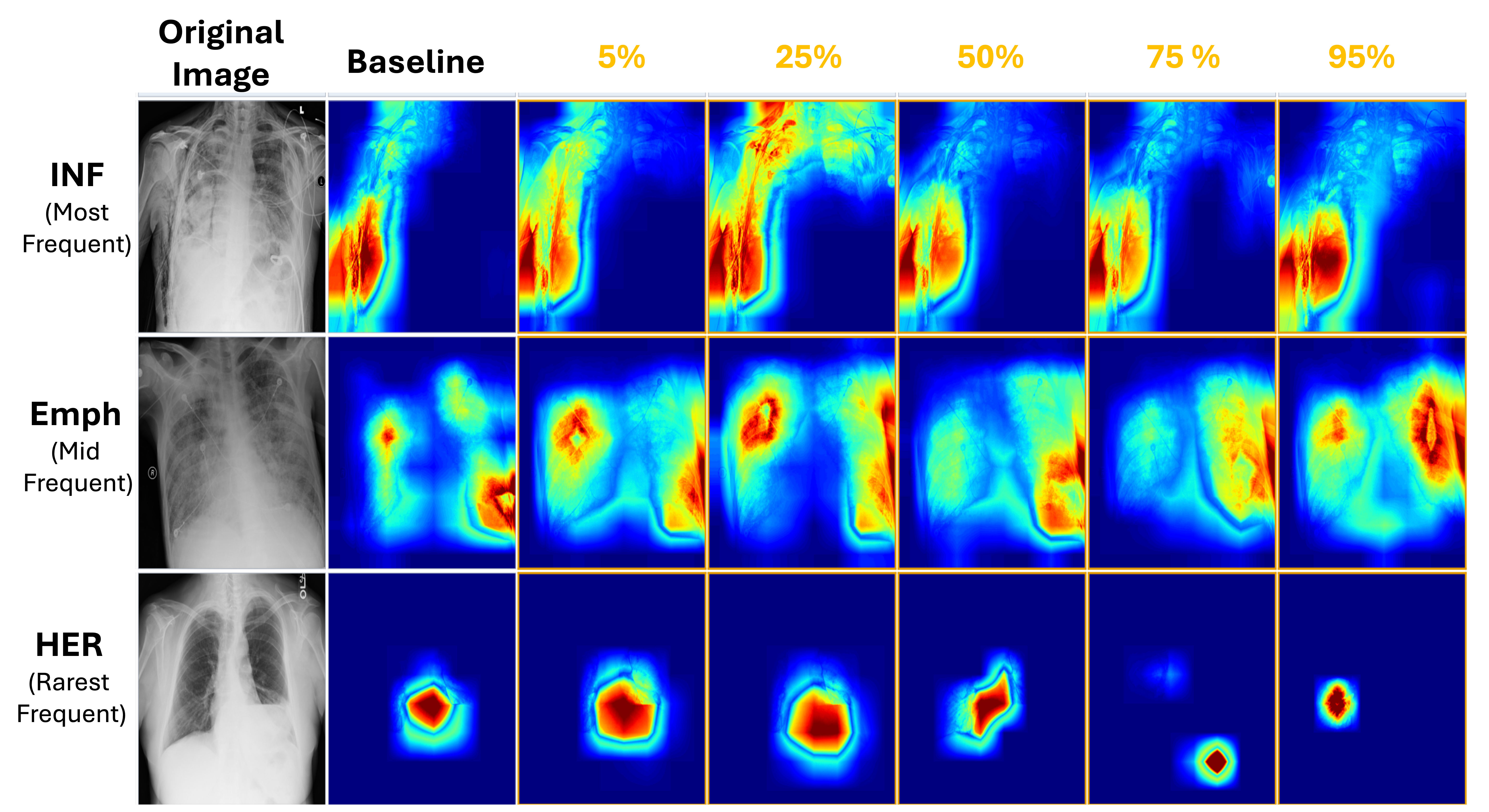}
  \caption{Effect of SNIP sparsity on Grad-CAM explanations. Grad-CAM maps for ResNet-50 on NIH-CXR-LT across increasing SNIP sparsity levels. Rows show representative head, mid, and tail classes: INF, EMPH, and HER, respectively. Grad-CAM localization remains largely consistent across increasing sparsity levels, with only gradual changes in the extent and intensity of the highlighted regions, illustrating the robustness of SNIP to aggressive pruning.}
    \label{fig:snip_gradcam}
\end{figure}

Fig.~\ref{fig:map_sparsity} summarizes the overall impact of pruning on predictive performance across four pruning strategies, two datasets, and two backbone architectures. Distinct compression behaviors emerge across the methods. SNIP exhibits the most graceful degradation overall, maintaining the highest or near-highest mAP across moderate-to-high sparsity levels. In contrast, L1 pruning remains competitive at low-to-moderate sparsity but exhibits a cliff-edge behaviour at high sparsity, where predictive performance deteriorates rapidly. GraSP incurs an immediate performance penalty following pruning but subsequently degrades more gradually, whereas Random pruning experiences severe degradation at low sparsity and remains at a low-performance plateau thereafter. These trends are remarkably consistent across datasets and architectures, with the only notable exception occurring on NIH-CXR-LT using ResNet-50 (Fig.~\ref{fig:map_sparsity}(a)), where L1 pruning slightly outperforms SNIP before undergoing its characteristic high-sparsity collapse.

While aggregate mAP provides an overall measure of compression performance, it conceals substantial class-dependent behavior. Fig.~\ref{fig:perclass_forgetting} illustrates the evolution of representative head, mid, and tail classes under increasing sparsity, whereas Figs.~\ref{fig:scatter_drop} and \ref{fig:scatter_rel} quantify the relationship between class frequency and pruning-induced degradation. Across all pruning methods, datasets, and architectures, a clear frequency-dependent trend emerges: higher-frequency classes generally exhibit greater robustness to pruning, whereas lower-frequency classes tend to experience earlier and larger performance degradation. Although individual classes deviate from this trend, lower-frequency classes generally reach a 20\% AP reduction at lower sparsity levels (Fig.~\ref{fig:scatter_drop}) and undergo larger degradation at 95\% sparsity (Fig.~\ref{fig:scatter_rel}). The consistently high Pearson and Spearman correlations across pruning methods, datasets, and architectures demonstrate a strong association between class frequency and pruning robustness.

While all pruning methods exhibit frequency-dependent degradation, their severity differs considerably. SNIP preserves class-wise performance most effectively and occasionally improves AP for several representative classes at low-to-moderate sparsity, possibly reflecting an implicit regularization effect. GraSP incurs an early performance penalty but maintains relatively stable degradation thereafter. In contrast, L1 pruning exhibits a pronounced high-sparsity collapse, with lower-frequency classes generally experiencing larger performance degradation than higher-frequency classes despite some class-specific exceptions. Random pruning performs worst, causing rapid degradation across all class-frequency groups even at low sparsity. These results demonstrate that gradient-informed pruning strategies preserve predictive performance more effectively across both aggregate and class-wise evaluations than magnitude-based and random pruning under aggressive compression.

\begin{figure*}[t]
    \centering
    \includegraphics[width=\textwidth]{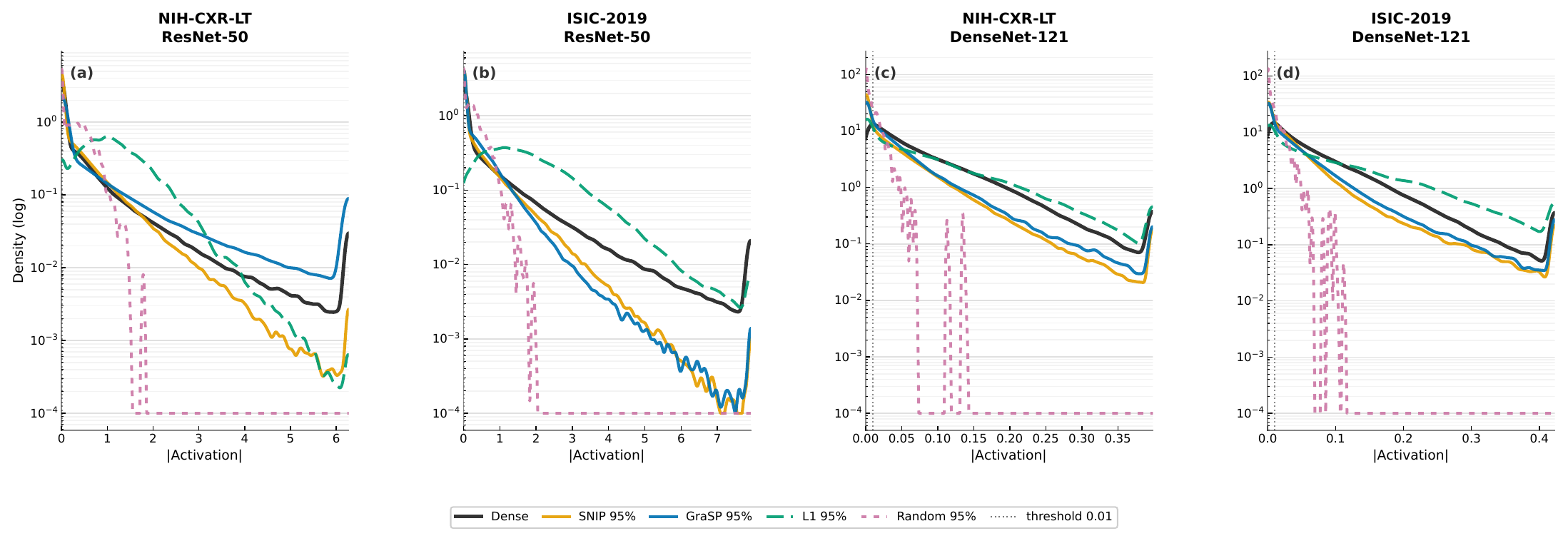}
    \caption{\textbf{Absolute activation-magnitude distributions under 95\%
pruning sparsity.} Each panel shows the estimated density (log scale) of
$|$Activation$|$ at the GradCAM target layer for the unpruned baseline
(Dense) and the 4 pruning techniques (SNIP, GraSP, L1, Random), on
NIH~=~NIH-CXR-LT and ISIC~=~ISIC-2019, with R50~=~ResNet-50 and
DN121~=~DenseNet-121. The vertical dotted line marks the near-zero
threshold ($0.01$); the exact share of near-zero activations per method
is reported in Table~\ref{tab:near_zero_summary}. Unlike the gradient
collapse in Fig.~\ref{fig:gradcam_gradient_ecdf}, activation magnitude does not
always shrink under pruning -- L1 pruning, for instance, leaves
activations comparatively intact even as its gradients vanish almost
completely, indicating units that still fire but no longer receive a
useful learning/attribution signal.}
    \label{fig:gradcam_activation_distributions}
\end{figure*}
\begin{figure*}[t]
    \centering
    \includegraphics[width=\textwidth]{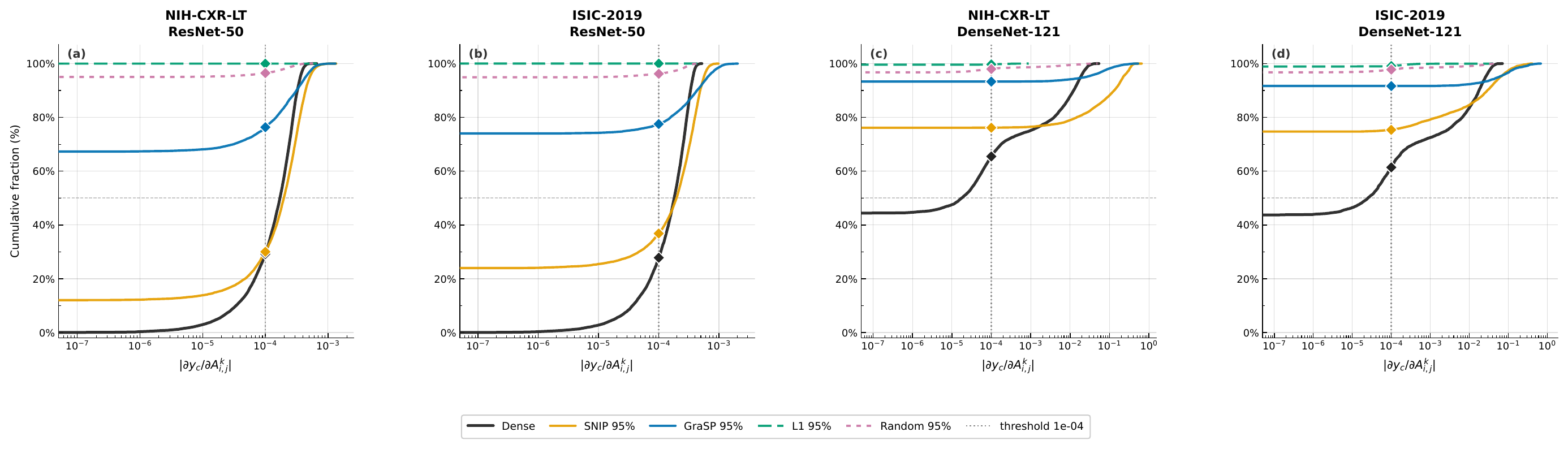}
    \caption{\textbf{GradCAM gradient sensitivity under 95\% pruning sparsity.}
Each panel shows the empirical CDF of $|\partial y_c/\partial A^k_{i,j}|$ --
the absolute gradient of the predicted class score w.r.t.\ the GradCAM
target-layer activations -- for the unpruned baseline (Dense) and the 4
pruning techniques (SNIP, GraSP, L1, Random), on NIH~=~NIH-CXR-LT and
ISIC~=~ISIC-2019, with R50~=~ResNet-50 and DN121~=~DenseNet-121. The
vertical dotted line marks the near-zero threshold ($10^{-4}$); diamonds
mark each curve's cumulative fraction at that threshold, with exact
percentages reported in Table~\ref{tab:near_zero_summary}. A curve sitting
further left/higher at the threshold means a larger share of gradients
has collapsed to numerically negligible values -- directly weakening the
signal GradCAM averages into its channel importance weights $\alpha_k$.}
    \label{fig:gradcam_gradient_ecdf}
\end{figure*}

\subsection{Explanation Stability Under Pruning}
\label{sec:results_stability}

Fig.~\ref{fig:stability_1x4} summarizes the evolution of explanation stability, measured using Stability IoU, across sparsity levels, while Fig.~\ref{fig:stability_4x4} illustrates the corresponding behavior for representative head, mid, and tail classes. Explanation stability depends strongly on the pruning strategy. L1 pruning preserves almost identical explanations to the dense model at low sparsity but undergoes a sharp collapse at high sparsity, closely mirroring the cliff-edge behavior observed for predictive performance. In contrast, SNIP exhibits the most gradual degradation, maintaining the highest spatial agreement with the dense model even at extreme sparsity. GraSP diverges from the dense model immediately after pruning but subsequently remains relatively stable across the remaining sparsity range, whereas Random pruning rapidly loses explanation consistency and maintains uniformly low Stability IoU throughout the remaining sparsity levels.

The per-class analysis in Fig.~\ref{fig:stability_4x4} shows that the effect of pruning on explanation stability is primarily method-dependent rather than consistently governed by class frequency. SNIP preserves relatively stable spatial correspondence across representative head-, mid-, and tail-frequency classes, while GraSP exhibits an early shift in explanation patterns followed by comparatively stable behavior. Although L1 pruning undergoes a pronounced collapse in explanation stability at high sparsity, this degradation is not consistently ordered by class frequency across datasets and architectures. Similarly, Random pruning rapidly degrades explanation stability across all representative classes, largely obscuring any frequency-dependent trends. In contrast to predictive performance, where long-tail forgetting is consistently observed, explanation stability does not exhibit a uniform head--mid--tail ordering. These results suggest that explanation stability is predominantly determined by the pruning strategy, with gradient-informed methods preserving spatial consistency substantially better than magnitude-based and random pruning under aggressive compression.

\subsection{Explanation Faithfulness Under Pruning}
\label{sec:results_aopc}

Fig.~\ref{fig:aopc_1x4} summarizes explanation faithfulness, measured using AOPC, across sparsity levels, while Fig.~\ref{fig:aopc_4x4} illustrates the corresponding behavior for representative head, mid, and tail classes. Compared with predictive performance, explanation faithfulness is substantially more robust to compression. SNIP and GraSP maintain relatively stable AOPC values across the entire sparsity range, indicating that the highlighted image regions continue to make substantial contributions to the model's predictions even under aggressive pruning. This behavior contrasts with the gradual decline in predictive performance, suggesting a partial decoupling between prediction accuracy and explanation faithfulness under compression.

The effect of pruning on explanation faithfulness is strongly influenced by the pruning strategy. L1 pruning preserves explanation faithfulness at low-to-moderate sparsity but undergoes a sharp collapse at high sparsity, closely matching the cliff-edge degradation observed for predictive performance and explanation stability. Random pruning performs worst, with AOPC approaching zero shortly after pruning begins, indicating that the identified salient regions no longer correspond to the model's decision process. The per-class analysis in Fig.~\ref{fig:aopc_4x4} does not exhibit a consistent head--mid--tail ordering, suggesting that differences in explanation faithfulness are primarily method-dependent rather than systematically associated with class frequency. Overall, gradient-informed pruning methods preserve explanation faithfulness substantially better than magnitude-based and random pruning under high compression.

\subsection{Qualitative and Mechanistic Analysis}
\label{sec:results_qualitative}

Fig.~\ref{fig:xai_pruning_comparison} presents qualitative Grad-CAM comparisons at 95\% sparsity across four pruning methods, two datasets, and two backbone architectures. Distinct explanation behaviors emerge across the pruning strategies. SNIP generally preserves lesion-centered attribution patterns that closely resemble those of the dense baseline, while GraSP maintains meaningful localization but exhibits broader and less focused activation regions. In contrast, L1 pruning undergoes pronounced degradation in explanation quality at extreme sparsity, producing diffuse or weak attribution maps. Random pruning performs worst, frequently generating diffuse, edge-dominated, or nearly blank attribution maps, indicating a substantial loss of class-discriminative localization. These qualitative trends are consistently observed across both datasets, and architectures, complementing the quantitative Stability IoU analysis.

To examine how explanations evolve with increasing compression, Fig.~\ref{fig:snip_gradcam} illustrates Grad-CAM maps for SNIP across sparsity levels. The highlighted regions evolve gradually with increasing sparsity, exhibiting moderate spatial drift and broader activation patterns while continuing to localize diagnostically relevant regions over a wide sparsity range. Even at high sparsity, clinically relevant structures remain identifiable, illustrating the robustness of gradient-informed pruning compared with the abrupt degradation observed under L1 pruning.

To better understand the mechanisms underlying explanation degradation, we first examined the activation distributions at the Grad-CAM target layer (Fig.~\ref{fig:gradcam_activation_distributions}; Table~\ref{tab:near_zero_summary}). While Random pruning substantially reduces activation magnitudes, L1 pruning retains substantial non-zero feature activations despite severe degradation in explanation stability and faithfulness, indicating that the presence of feature activations alone is insufficient to preserve reliable explanations.

We therefore analyzed the class-discriminative gradients used to compute Grad-CAM (Fig.~\ref{fig:gradcam_gradient_ecdf}; Table~\ref{tab:near_zero_summary}). L1 and Random pruning shift the gradient distribution toward near-zero values, with the vast majority of gradients falling below the selected threshold, whereas SNIP retains substantially stronger gradient signals across both datasets and backbone architectures. Since Grad-CAM directly relies on these gradients to weight feature maps, the collapse of gradient magnitudes provides a mechanistic explanation for the degenerate attribution maps observed under extreme sparsity. Hence, Figs.~\ref{fig:gradcam_activation_distributions}, \ref{fig:gradcam_gradient_ecdf}; Table~\ref{tab:near_zero_summary} indicate that preserving feature activations alone is insufficient to maintain reliable Grad-CAM explanations. Instead, explanation quality is strongly associated with preserving class-discriminative gradients, providing a plausible mechanistic explanation for the degradation observed under aggressive pruning.

\begin{table*}[t]
\centering
\caption{\textbf{Near-Zero Gradient and Activation Fractions Under 95\% Pruning Sparsity.} NIH~=~NIH-CXR-LT, ISIC~=~ISIC-2019, R50~=~ResNet-50, DN121~=~DenseNet-121. \textbf{Grad\,\%}: share of GradCAM target-layer gradients $|\partial y_c/\partial A^k_{i,j}|$ below $10^{-4}$. \textbf{Act\,\%}: share of $|$Activation$|$ values below $0.01$. Both reported at 95\% sparsity, except Dense (unpruned baseline); higher values indicate more of the signal has collapsed to dead, uninformative values, directly degrading GradCAM's saliency computation. L1 pruning drives gradients to near-total collapse (99--100\%) while activations remain comparatively intact; Random pruning collapses both, with activations far sparser on DN121 than R50.}
\label{tab:near_zero_summary}
\scriptsize
\setlength{\tabcolsep}{4pt}
\begin{tabular*}{\linewidth}{@{\extracolsep{\fill}} l cccccccc}
\toprule
Method & \multicolumn{2}{c}{NIH R50} & \multicolumn{2}{c}{ISIC R50} & \multicolumn{2}{c}{NIH DN121} & \multicolumn{2}{c}{ISIC DN121} \\
\cmidrule(lr){2-3} \cmidrule(lr){4-5} \cmidrule(lr){6-7} \cmidrule(lr){8-9}
 & Grad \% & Act \% & Grad \% & Act \% & Grad \% & Act \% & Grad \% & Act \% \\
\midrule
Dense & 29.2 & 58.4 & 27.8 & 51.1 & 65.5 & 14.9 & 61.4 & 17.9 \\
SNIP & 30.0 & 48.7 & 36.8 & 49.4 & 76.2 & 51.1 & 75.4 & 43.9 \\
GraSP & 76.3 & 55.7 & 77.5 & 40.5 & 93.3 & 41.7 & 91.6 & 44.1 \\
L1 & 100.0 & 3.4 & 100.0 & 1.6 & 99.7 & 27.8 & 99.1 & 25.7 \\
Random & 96.5 & 32.1 & 96.2 & 23.4 & 97.9 & 67.2 & 97.8 & 68.6 \\
\bottomrule
\end{tabular*}
\end{table*}

\section{Discussion}
\label{sec:discussion}

\subsection{Why Does Pruning Disproportionately Affect Rare Diseases?}

A central finding of this work is that long-tail forgetting is a systematic consequence of model pruning rather than a property of a particular pruning algorithm. Across all datasets, architectures, and pruning methods, pruning robustness exhibits a strong association with class frequency. Although individual classes deviate from this trend, lower-frequency classes generally experience earlier and larger performance degradation than higher-frequency classes. We hypothesize that this behavior arises from differences in representational redundancy. Frequent diseases are observed repeatedly during training and therefore develop richer and more redundant feature representations, whereas rare diseases rely on comparatively fewer discriminative pathways. Consequently, globally applied pruning criteria are more likely to remove parameters that are critical for rare-class recognition.

Although long-tail forgetting is consistently observed, its severity varies across pruning methods. Gradient-informed approaches, particularly SNIP, preserve class-wise performance substantially better than magnitude-based and random pruning, suggesting that retaining optimization-sensitive connections is beneficial for maintaining minority-class representations under aggressive compression. These observations extend previous studies that primarily investigated magnitude-based pruning by demonstrating that long-tail forgetting generalizes across multiple pruning strategies.

\subsection{Explanation Stability and Faithfulness Provide Complementary Insights}

Explanation stability and explanation faithfulness exhibit distinct responses to model pruning, indicating that they capture complementary aspects of explanation reliability. Unlike predictive performance, neither metric exhibits a consistent frequency-dependent ordering across representative classes. Instead, both explanation stability and explanation faithfulness are influenced primarily by the pruning strategy, suggesting that explanation robustness and predictive robustness respond differently to compression. While explanation stability decreases progressively with increasing sparsity, explanation faithfulness remains comparatively robust for gradient-informed pruning methods over a wide range of sparsity levels. This indicates that compressed models may continue to rely on clinically meaningful image evidence even when predictive performance begins to decline.

These observations further highlight that explanation stability and explanation faithfulness quantify different properties of model behavior. Stability measures the consistency of explanations relative to the dense model, whereas faithfulness evaluates whether highlighted regions causally contribute to the model's prediction. Consequently, a pruned model may generate explanations that differ spatially from those of the dense model while still highlighting image regions that remain important for its predictions. Evaluating both metrics therefore provides a more comprehensive assessment of explanation reliability than either metric alone.

\subsection{Mechanisms Underlying Explanation Degradation}

The qualitative Grad-CAM analysis, together with the activation and gradient studies, provides insight into why explanations degrade under aggressive pruning. While Random pruning substantially weakens feature activations, L1 pruning largely preserves activation magnitudes despite producing severely degraded attribution maps. In contrast, both methods exhibit a pronounced collapse of class-discriminative gradients at the Grad-CAM target layer. Since Grad-CAM directly relies on these gradients to weight feature maps, the collapse of gradient information provides a plausible explanation for the degenerate attribution maps observed under extreme sparsity.

These findings suggest that preserving feature activations alone is insufficient to maintain reliable explanations after aggressive pruning. Instead, explanation quality depends critically on preserving class-discriminative gradients. This may also help explain the favorable behavior of SNIP, which retains connections with high first-order sensitivity to the training loss, although its objective does not explicitly optimize explanation gradients. More broadly, these results demonstrate that preserving predictive performance alone is insufficient to ensure reliable explanations after compression.

\subsection{Implications for Deployment of Compressed Medical AI}

Our findings have several implications for the deployment of compressed medical AI systems. First, aggregate performance metrics alone are insufficient because they can obscure substantial degradation in clinically important but infrequent diseases. Per-class evaluation is therefore essential, particularly for long-tailed medical datasets where aggregate metrics may conceal clinically important failure modes. Second, explanation reliability should be evaluated alongside predictive performance, as compression affects these properties differently depending on the pruning strategy. Finally, moderate sparsity levels consistently provide the best compromise between compression, predictive performance, and explanation reliability across datasets and architectures.

These findings suggest that compressed medical AI systems should be evaluated using class-aware and explanation-aware protocols rather than aggregate performance alone. Such evaluations provide a more realistic assessment of model robustness in safety-critical clinical applications, where failures on rare diseases or unreliable explanations may have important clinical consequences.

\section{Conclusion}
\label{sec:conclusion}

This paper presents a systematic study of long-tail forgetting and explanation reliability under model pruning in medical imaging. Across four pruning strategies, two architectures, and two long-tailed datasets, we show that long-tail forgetting is consistently observed under pruning, with lower-frequency classes degrading earlier and more severely than higher-frequency classes. We further demonstrate that pruning affects explanation reliability in a method-dependent manner, with gradient-informed pruning, particularly SNIP, providing the most favorable balance between predictive performance and explanation reliability. Our qualitative and mechanistic analyses suggest that degradation of Grad-CAM explanations under aggressive pruning is primarily associated with the collapse of class-discriminative gradients rather than the disappearance of high-level feature activations. These findings highlight the importance of evaluating compressed medical AI models using class-aware and explanation-aware protocols in addition to conventional aggregate performance metrics. Moderate sparsity levels consistently provide a practical operating regime that preserves both predictive performance and explanation reliability.

While this study focuses on CNN architectures, extending the analysis to transformer-based models will be important to determine whether the observed pruning behaviors generalize across different architectural paradigms. Future work will also investigate other compression strategies, including structured pruning, quantization, and knowledge distillation, and broader medical imaging tasks such as localization, detection, and segmentation.

\section*{References}

\end{document}